\documentclass[10pt,twocolumn,letterpaper]{article}

\usepackage[pagenumbers]{cvpr} 

\usepackage{graphicx}
\usepackage{amsmath}
\usepackage{amssymb}
\usepackage{booktabs}
\usepackage{nicefrac}
\usepackage{xcolor,colortbl}
\usepackage{multirow}
\usepackage{sidecap}

\newcommand{\red}[1]{\textcolor{red}{#1}}

\definecolor{vvlightgray}{rgb}{0.9,0.9,0.9}
\definecolor{vlightgray}{rgb}{0.8,0.8,0.8}

\usepackage[pagebackref,breaklinks,colorlinks]{hyperref}

\usepackage[capitalize]{cleveref}
\crefname{section}{Sec.}{Secs.}
\Crefname{section}{Section}{Sections}
\Crefname{table}{Table}{Tables}
\crefname{table}{Tab.}{Tabs.}

\def\cvprPaperID{7099} 
\def\confName{CVPR}
\def\confYear{2022}

\begin{document}

\title{Integrating Language Guidance into Vision-based Deep Metric Learning}

\author{Karsten Roth$^1$, Oriol Vinyals$^2$, Zeynep Akata$^{1, 3}$\\
$^1$University of Tübingen, $^2$DeepMind, $^3$MPI for Intelligent Systems}
\maketitle

\begin{abstract}
Deep Metric Learning (DML) 
proposes to learn metric spaces which encode semantic similarities as embedding space distances. These spaces should be transferable to classes beyond those seen during training. 
Commonly, DML methods task networks to solve contrastive ranking tasks defined over binary class assignments.
However, such approaches ignore higher-level semantic relations between the actual classes. This causes learned embedding spaces to encode incomplete semantic context and misrepresent the semantic relation between classes, impacting the generalizability of the learned metric space.
To tackle this issue, we propose a language guidance objective for visual similarity learning. Leveraging language embeddings of expert- and pseudo-classnames, we contextualize and realign visual representation spaces corresponding to meaningful language semantics for better semantic consistency.
Extensive experiments and ablations provide a strong motivation for our proposed approach and show language guidance offering significant, model-agnostic improvements for DML, achieving competitive and state-of-the-art results on all benchmarks. Code available at \href{https://github.com/ExplainableML/LanguageGuidance_for_DML}{github.com/ExplainableML/LanguageGuidance\_for\_DML}.
\end{abstract}

\section{Introduction}
\label{sec:intro}
\begin{figure}[t]
    \centering
    \includegraphics[width=0.47\textwidth]{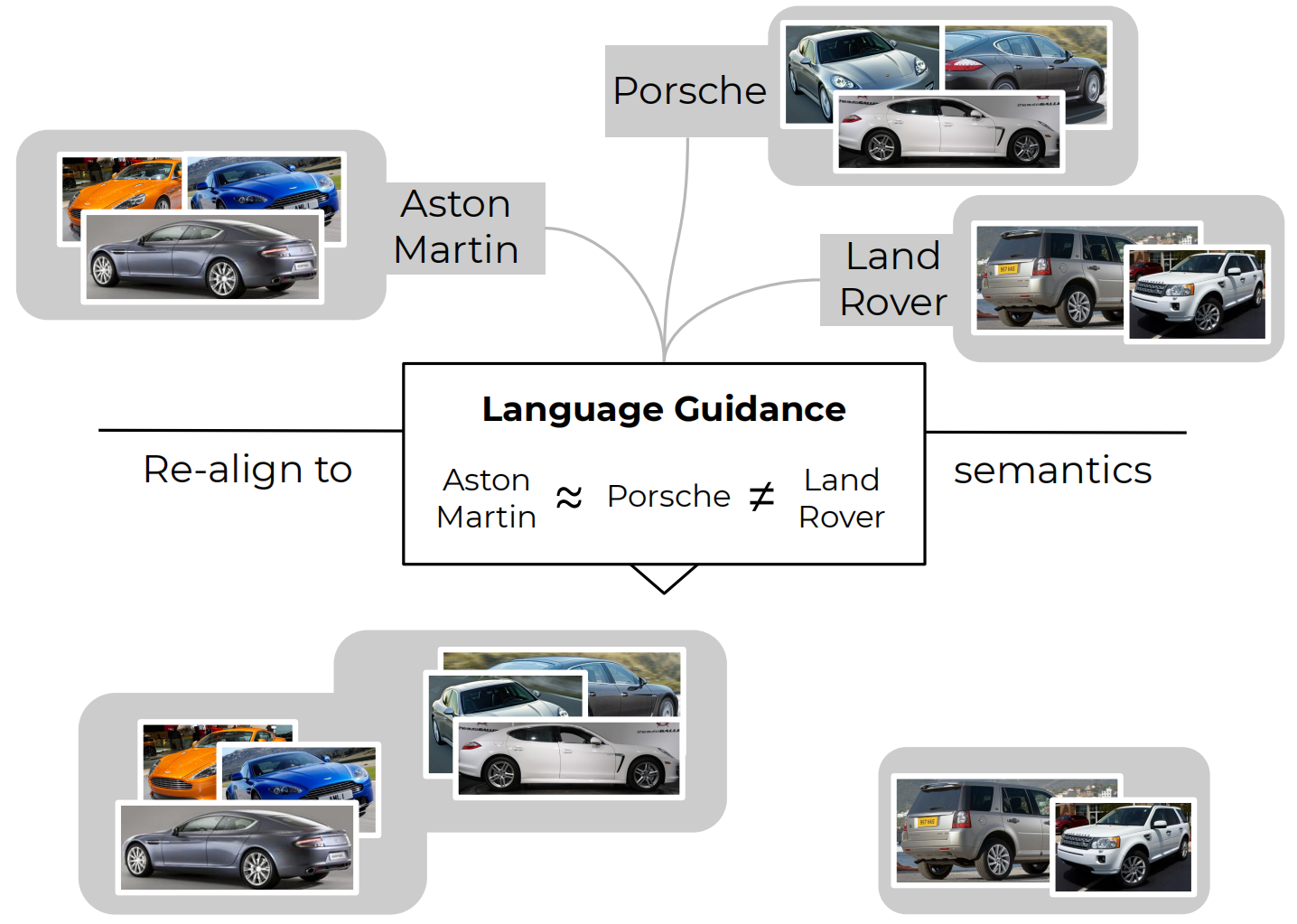}
    \vspace{-6pt}
    \caption{\textit{Language-guidance for better semantic visual alignment.} We leverage language context to better align learned metric spaces with higher-level semantic relations and significantly improve generalization performance.
    }
    \label{fig:firstpage}
    \vspace{-8pt}
\end{figure}
Visual similarity learning with deep networks drives important applications such as image retrieval \cite{semihard,margin}, face verification \cite{sphereface,arcface}, clustering \cite{grouping} or contrastive supervised \cite{khosla2020supervised} and unsupervised representation learning \cite{moco,chen2020simple}.
Deep Metric Learning (DML) has proven a useful and widely adopted framework to contextualize visual similarities by learning (deep) metric representation/embedding spaces in which a predefined distance metric, such as the euclidean or cosine distance, has a strong connection to the actual underlying semantic similarity of two samples \cite{roth2020revisiting,musgrave2020metric}.

In most visual similarity tasks, transfer beyond just the training distribution and classes is crucial, which requires learned representation spaces to encode meaningful semantic context that generalizes beyond relations seen during training.
However, the majority of DML methods introduce training paradigms based only around class labels provided in given datasets to define ranking tasks for networks to solve. 
This treats every class the same, with the arrangement of classes in embedding space solely derived from the class-label generated ranking tasks. 
In doing so, high-level semantic connections between different classes (e.g. sports cars vs. pickup trucks) can't be accounted for, even though semantic context beyond what can be derived from purely discriminative class labels facilitates stronger generalization especially to novel classes \cite{milbich2020diva,dvml,hardness-aware,mic}.
And while contextualization via e.g. the definition of hierarchies \cite{word_hierarchies,hier_lit_1,hier_lit_2,hier_lit_3} can help refine classlabels, such approaches commonly rely on predefined rules or expert knowledge.

To address this problem, we propose to leverage the large corpora of readily available, large-scale pretrained natural language models \footnote{S.a. transformer language models \cite{transformers} with strong zero-/fewshot generalization \cite{gpt2,gpt3,clip,bert,roberta} across a large variety of applications. Provided e.g. through public libraries such as \texttt{huggingface} \cite{huggingface}.} to provide task-independent contextualization for class labels and encourage DML models to learn semantically more consistent visual representation spaces.\\
Using language-based pretraining for contextualization of visual similarities is long overdue - for vision-based DML, pretraining (on ImageNet) has already become standard \cite{semihard,margin,arcface,abier,Sanakoyeu_2019_CVPR,roth2020revisiting,musgrave2020metric,milbich2020diva,seidenschwarz2021graphdml} since it provides a strong and readily available starting point. 
This starting point transfers well to a multitude of downstream domains and ensures ranking tasks underlying most DML methods to be much better defined initially, improving training and generalization performance (e.g. \cite{roth2020revisiting,musgrave2020metric}). 
Such ImageNet pretraining is standard, and crucial, even for unsupervised DML \cite{unsup_1,unsup_2,unsup_3,unsup_4}, and can similar be found in other areas of Deep Learning such as image detection \cite{detect_1,detect_2,detect_3,detect_4,detect_5,detect_6,detect_7}.
Thus, there is little reason to bottleneck DML to only leverage visual pretraining while disregarding the potential benefits of language context.

To incorporate pretrained language models to facilitate visual similarity learning, we therefore propose \textit{language guidance} \textit{(ELG)} for DML. Given natural language class names, language embeddings and respective language similarities are computed, which are then used via distillation to re-arrange and correct visual embedding relations learned by standard DML methods.
However, natural language class names require expert knowledge. To circumvent the need for such additional supervision, we further propose \textit{pseudolabel-guidance} \textit{(PLG)}. 
Leveraging the ubiquitously used ImageNet pretraining in DML pipelines, we define a quasi-unique collection of natural language ImageNet pseudolabels for samples and classes essentially ``for free''. Re-embedding these pseudolabels into pretrained language models gives access to a collection of less finegrained, but generically applicable pseudolabel similarities which can then similarly be used for language guidance with little changes in generalization performance. 
Extensive experimentation and ablations support the validity of our proposed approach and showcase significant improvements to the generalization performance of DML models when using pretrained language models for additional visual semantic refinement, concurrently setting a new state-of-the-art with negligible overhead to training time.

\section{Related Works}
\begin{figure*}[t]
    \centering
    \includegraphics[width=0.97\textwidth]{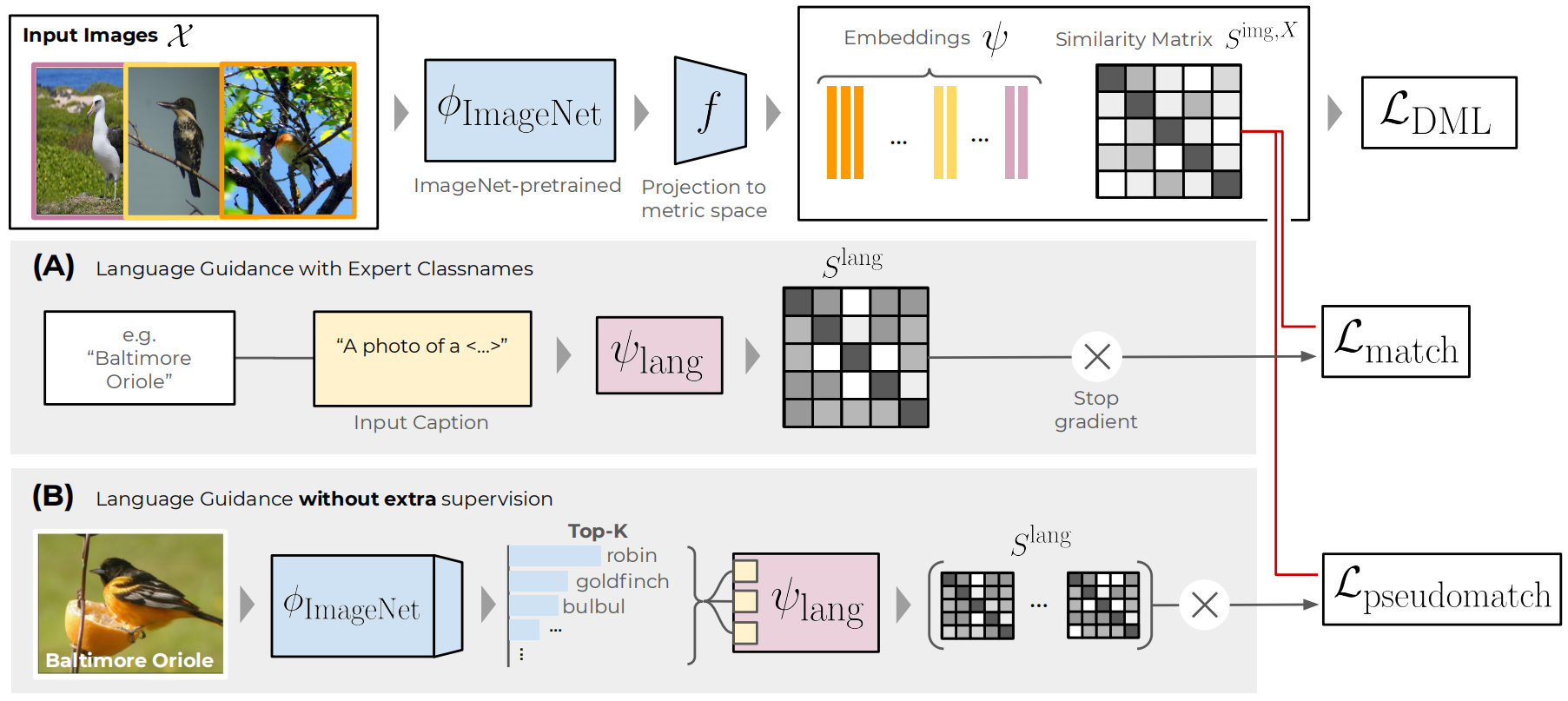}
    \vspace{-6pt}
    \caption{\textit{Language guidance.} We extend the default DML pipeline for Visual Similarity Learning by embedding either \textbf{(A)} expert class names or \textbf{(B)} top-$k$ ImageNet pseudolabels, which require no additional expert supervision, with a pretrained language model. This provides language similarity matrices $S^\text{lang}$ which are used to guide the structuring of our finegrained visual similarity space generated by $f\circ \phi_\text{ImageNet}$ through distillation-based matching ($\mathcal{L}_\text{match}$ \& $\mathcal{L}_\text{pseudomatch}$) between $S^\text{lang}$ and image similarities $S^\text{img, X}$.
    }
    \label{fig:setup}
    \vspace{-5pt}
\end{figure*}
\label{sec:related_works}

\textbf{Deep Metric Learning.} 
Visual Similarity Learning driven by advances in Deep Metric Learning has proven essential for zero-shot applications in image \cite{semihard,margin,mic,Sanakoyeu_2019_CVPR} and video retrieval \cite{Brattoli_2020_CVPR,Wray_2021_CVPR}, clustering \cite{margin,htl,sohn2018unsupervised,Yan_2021_CVPR} or person re-identification \cite{sphereface,arcface}, but also for unsupervised and supervised representation learning methods relying on contrastive training \cite{moco,chen2020simple,pretextmisra,tian2019contrastive}.
Research in Deep Metric Learning can be separated into various conceptual directions, such as the development of training surrogates via ranking losses, e.g. through pairwise \cite{contrastive}, triplet \cite{face_verfication_inthewild}, quadruplet \cite{quadtruplet}, higher-order ranking constraints \cite{lifted,npairs,multisimilarity} or graph-based reweighting \cite{Zhu2020graphdml,seidenschwarz2021graphdml}, but also using regularizations and constraints on the incorporated distance functions \cite{tian2020hynet,horde}. 
However, as the usage of tuples introduces an exponential increase in sample complexity \cite{semihard,margin}, similar emphasis has also been placed in tuple selection heuristics to boost training speeds and generalization, either based on sample distances \cite{semihard,margin,multisimilarity,Suh_2019_CVPR}, hierarchical arrangements \cite{htl} or adapted to the training process \cite{smartmining,roth2020pads}.
Tuple complexity can also be addressed using proxies as stand-in replacements in the generation of tuples \cite{proxynca,softriple,kim2020proxy,proxyncapp,Zhu2020graphdml,Deng_2021_CVPR}.
However, while literature results suggests increasing generalization performance based on simple changes in re-ranking and tuple selection, recent work has instead highlighted a much stronger saturation in method performance \cite{fehervari2019unbiased,roth2020revisiting,musgrave2020metric}, underlining the importance of fair and comparable training and evaluation protocols with fixed backbone network and pipeline parameter choices. For our experimental evaluation, we make sure to provide a thorough and fair comparison to allow for transferable insights.
Fortunately, recent work have shown the benefit of orthogonal extensions to the common DML training approaches with strong relative improvements, such as through auxiliary feature learning \cite{mic,milbich2020sharing,milbich2020diva,dullerud2022is}, artificial data generation \cite{hardness-aware}, adversarial and variational extensions \cite{daml,dvml}, virtual class context \cite{ko2021learning}, solving DML in metric subspaces \cite{Sanakoyeu_2019_CVPR,abe,dreml,abier}, knowledge distillation and transfer \cite{darkrank,s2sd,Kim_2021_CVPR}, distance regularization \cite{Mohan_2020_CVPR}, mutual learning \cite{diversifieddml} or few-shot adaptation \cite{milbich2021characterizing}. Our research into language-guided deep metric learning follows this line of thought and aims to extend the generalizability of learned metric spaces through the usage of dense language context incorporate beneficial semantics without the need for additional explicit supervision.

\textbf{Crossmodal Similarity Learning.} 
Our proposed approach, through the usage of implicit language supervision, ties into related works in crossmodal similarity learning, which finds applications in cross-modal retrieval and representation learning \cite{crossmodal_1,crossmodal_2,crossmodal_3,crossmodal_4, crossmodal_5, crossmodal_6, tsimpoukelli2021multimodal,Dzabraev_2021_CVPR,crossmodal_7,crossmodal_8,perceiver}, where joint training can also support the generalization capabilities in single modalities, such as shown in \cite{devise,clip,virtex}. 
For this work, our focus was placed primarily on language as a secondary modality to guide visual similarity learning. While hierarchical knowledge graphs (such as via WordNet \cite{wordnet} or more specific taxonomies) and color name priors have seen usage in representation learning tasks \cite{cm_1,cm_2,hier_cm_1,hiermatch,hier_lit_1,hier_lit_2,hier_lit_3,hier_lit_4,hier_lit_5} and single-noun prior clustering \cite{cohen2021singlenoun}, we propose to leverage dense relations between language representations generated by large, zero-shot capable language models \cite{gpt2,gpt3,clip} to re-arrange finegrained visual similarity spaces through relative matching (unlike e.g. language embedding prediction for image classification in \cite{devise}.
In doing so, we find significant improvements in generalization performance across benchmarks while circumventing the need for task-specific semantic hierarchies and rules. In addition to that, we also show how our language guidance can be applied in settings where no expert class information is available.

\section{Language Guidance}
\label{sec:methods}
This section addresses DML preliminaries, language guidance with expert class labels (\S\ref{subsec:language_context}) and language guidance without additional supervision (\S\ref{subsec:language_guided}).

\subsection{Preliminaries}
\label{subsec:preliminaries}
Deep Metric Learning (DML) learns a distance metric $d_\psi(x_1, x_2)$ over images $x_i\in\mathcal{X}$ parametrized by a deep feature extraction model $\phi:\mathcal{X}\rightarrow \Phi$ (in contrast to handcrafted feature extractors in setups predating the advent of Deep Learning, see e.g \cite{surez2018tutorial}) and a projection to the target metric space $f: \Phi\rightarrow \Psi\subset\mathbb{R}^d$, which defines a Mahalanobis (pseudo-)distance over features \cite{surez2018tutorial}. 
While non-end-to-end trainable methods primarily optimize for a parametrized metric over given features, in DML, both are trained jointly. 
This allows us to learn a projection from image space, $\psi = f \circ \phi$, that spans a metric (or embedding) space $\Psi$ such that a predefined distance metric $d(\psi_1, \psi_2)$ on $\Psi$, usually the cosine or euclidean distance $d = \left\Vert\bullet, \bullet\right\Vert$, has a close connection to the true semantic similarity of input samples. 
$\Psi$ is commonly normalized to the unit hypersphere $\Psi := \mathcal{S}_\Psi$ \cite{margin,arcface,multisimilarity,milbich2020diva,roth2020revisiting} for regularization \cite{hypersphere,margin,wang2020understanding,roth2020revisiting}.
Such high-dimensional embedding spaces suitable for use with easy predefined, non-parametric metrics are attractive for fast, approximate similarity search methods \cite{faiss,ann_1,ann_2}.
In supervised DML, $\Psi$ is commonly learned with ranking tasks defined using provided class label information, introducing contrastive objectives based on pairs, triplets or tuples of higher order. 
Taking for example pairs $(x_a, x_p)$ or $(x_a, x_n)$ where $y_a = y_p \neq y_n$ with \textit{anchor} $x_a$, \textit{positive} $x_p$ and \textit{negative} $x_n$, one can define a training objective following e.g. \cite{contrastive} and \cite{musgrave2020metric} as
\begin{equation}
\begin{split}
    \mathcal{L} &= \frac{1}{\mathcal{P}_\mathcal{B}} \sum_{(x_1, x_2) \in \mathcal{P}_\mathcal{B}} \mathbb{I}_{y_1 = y_2} \max[\gamma_p, d_\psi(x_1, x_2)]\\ &- \mathbb{I}_{y_1 \neq y_2} \min[\gamma_n, d_\psi(x_1, x_2)]
\end{split}
\end{equation}
with valid pairs $\mathcal{P}_\mathcal{B}$ in minibatch $\mathcal{B}$.
Doing so embeds same-class samples closer (according to $d(\bullet, \bullet)$) while pushing different classes apart up to margins $\gamma_p$ and $\gamma_n$. This can be easily extended to incorporate more complex relations by choice of higher-order tuples, tuple sampling heuristics \cite{semihard,margin,roth2020pads} or proxy-representations \cite{proxynca,proxyncapp,kim2020proxy}.

\subsection{Language Guidance with Expert Class Names}
\label{subsec:language_context}
The reliance on ranking tasks defined solely using class labels however does not address high-level semantic relations between classes, even though such non-discriminative relations are crucial for strong downstream generalization \cite{milbich2020diva,dvml,hardness-aware,mic}.
As such, we propose to leverage language semantics to better align visual representation spaces, and do so in two ways.
This section will introduce the \textit{E}xpert \textit{L}anguage \textit{G}uidance (\textit{ELG}) approach (see also Fig. \ref{fig:setup}a), which makes uses of expert class label names, while the next section covers \textit{P}seudolabel \textit{L}anguage \textit{G}uidance (\textit{PLG}).

To incorporate language semantics into visual similarity learning via \textit{ELG}, we make use of large pretrained language models $\psi_\text{lang}$ like CLIP \cite{clip}, BERT \cite{bert} or RoBERTa \cite{roberta}, which map an input sentence $c_i\in\mathcal{C}$, corresponding to the images $x_i\in\mathcal{X}$ ground truth class label, to the respective language embedding space $\Psi_\text{lang}$. For brevity, we use $\psi_i^\text{lang} := \psi_\text{lang}(c_i)$.
Given the minibatch of images $\mathcal{B}_\mathcal{X}\subset\mathcal{X}$ and labels $\mathcal{B}_\mathcal{Y}\subset\mathcal{Y}$, we generate input sequences $\mathcal{B}_\mathcal{C}\subset\mathcal{C}$ for the natural language model using some primer, e.g. $c_i = $ \texttt{"A photo of a }$y_i$\texttt{"}\footnote{Special characters are adjusted to match the expected model input (e.g. \texttt{027.Shiny\_Cowbird} $\rightarrow$ \texttt{Shiny Cowbird}.}. 
Let $\mathcal{B}_{\Psi_\text{lang}}$ be the natural language representation generated by $\psi_\text{lang}$ from $\mathcal{B}_\mathcal{C}$ and $S^\text{lang}$ the corresponding batch-wise cosine similarity matrix for all language representation in the minibatch. Similarly, let $S^\text{img}$ be the corresponding visual similarity matrix for $\mathcal{B}$. We then define the language distillation loss
\begin{equation}\label{eq:base_match}
\begin{split}
    \mathcal{L}_\text{match}&\left(\mathcal{S}^\text{img}, \mathcal{S}^\text{lang}\right) = \\ &\frac{1}{|\mathcal{B}|}\sum_{i}^{|B|}\sigma\left(S^{\text{img}, X}_{i}\right)\log\left(\frac{\sigma\left(S^{\text{img}, X}_{i}\right)}{\sigma\left(S^\text{lang}_{i} + \gamma_\text{lang}\right)}\right)
\end{split}
\end{equation}
as KL-Divergence matching between row-wise image- and language-level similarities $S^{\text{img}, X}_i$ and $S^\text{lang}_i$, similar to contrastive distillation objectives utilized in e.g. \cite{tian2019contrastive,s2sd}. 
Here $\sigma(\mathbf{x}_i) = \nicefrac{\exp(\mathbf{x}_i)}{\sum_j^{|\mathbf{x}|} \exp(\mathbf{x}_j)}$ denotes a row-wise softmax with shift $\gamma_\text{lang}$ to address the fidelity of the distribution matching between image and language similarity distributions.
Finally, $S^\text{lang}$ does not resolve similarities for samples within a class unlike $S^\text{img}$ (for $x_i \neq x_j$ and $y_i = y_j$ we have $\psi_i \neq \psi_j$ and thus $S^\text{img}_{i,j} < 1$, but, since the classes are the same, $S^\text{lang} = 1$). To ensure that we do not lose intraclass resolution during distillation, we thus adapt $S^\text{img}$:
\begin{equation}\label{eq:base_adapted}
S^{\text{img}, X}_{i, j} = 
\mathbb{I}_{y_i=y_j}\left[1 + \gamma_\text{lang}\right] + \mathbb{I}_{y_i\neq y_j}\left[S^\text{img}_{i, j}\right]
\end{equation}
which masks out respective row entries of $S^\text{img}$ where the classes are the same as the respective anchor class (plus the respective language offset $\gamma_\text{lang}$. This ignores matching within-class similarities for datasets where no sample-specific natural language description is available. Note that during training, backpropagation only occurs through $S^\text{img}$, as $S^\text{lang}$ only provides the language target which the image representation space should be aligned towards. For the remainder of this work, we use "+\textit{ELG}" to denote objectives that have been augmented with language guidance:
\begin{equation}\label{eq:final_eq}
    \mathcal{L}_\textit{ELG} = \mathcal{L}_\text{DML} + \omega \cdot \mathcal{L}_\text{match}
\end{equation}

\subsection{Language Guidance without extra supervision}
\label{subsec:language_guided}
While \textit{ELG} incorporates language context for visual similarity learning well, it requires expert class labelling. Fortunately, the ImageNet pretrained backbone (used in every DML pipeline) provides a suitable work-around. While previous work relies solely on the pretrained features to provide a starting point for the support over which the metric space is spanned from, we go a step further and also leverage the classifier head to produce ImageNet pseudolabels.
We then run both pretrained backbone and classifier head over all images within a class. For each sample, this allows us to produce softmax outputs corresponding to all ImageNet classes. These are then averaged for each class, and the top-$k$ ImageNet pseudo-classnames $\mathcal{Y}^{\text{IN}, k}_i$ are selected to represent the class (see Fig. \ref{fig:setup}\textbf{(B)} for examples). 

While the pseudo classlabels are not as finegrained as expert labels, it is fair to assume that the generic object recognized in the target image $x_i$ has, for the majority of cases, some relation with the true label $y_i$ (e.g. \texttt{sports car} instead of \texttt{Aston Martin Coupe}). Running the language network $\psi_\text{lang}$ on the pseudolabels then provides approximate semantic context. 
By repeating this process for all $k$ of the top-$k$ ImageNet-labels for each class, a more finegrained language resolution is then achieved, with the collection of top-$k$ ImageNet-labels providing a much more unique semantic description than a single pseudolabel does.
We then compute a collection of language-based similarity matrices $\left\{S^{\text{pseudolang}, i}\right\}_{i\in[0, ..., k-1]}$, which respects the ordering of pseudolabels, i.e. $S^{\text{pseudolang}, 1}$ denotes the similarity matrix between language embeddings generated by embedding the respective highest matching pseudolabels, which we found to work best in practice (c.f. \S\ref{subsec:conceptual_ablations}).

These language similarities are embedded into training by extending the matching objective in Eq. \ref{eq:base_match} to
\begin{align}
    \mathcal{L}_\text{pseudomatch}^k &= \mathcal{L}_\text{match}\left(S^\text{img}, \frac{1}{k}\sum_j^k S^\text{pseudolang, j}\right)\label{eq:multi_match}
\end{align}
by merging all $S^{\text{pseudolang}}$ (Eq. \ref{eq:multi_match}).
In doing so, image representation are aligned to the semantics of multiple class concepts closely related to the underlying image. 
In addition, unlike matching in \textit{ELG} (Eq. \ref{eq:base_match}), we can now also address semantic differences between images of the same class,
as pseudolabels can be extracted on a sample- instead of just the class level. 
This can be done by simply utilizing the unmasked image similarities $S^\text{img}$ instead of its masked variant $S^{\text{img}, X}$ (see Eq. \ref{eq:base_match} and \ref{eq:base_adapted}).
However, in practice we found no improvements on the evaluated benchmarks (cf. \S\ref{subsec:conceptual_ablations}). 
For the remainder of this work, we mark language guidance using ImageNet-pseudolabels with "+\textit{PLG}".

\begin{table*}[t]
    \caption{\textit{State-of-the-art.} \textbf{Bold}: best results per literature setup. The results showcase competitive and state-of-the-art performance with little hyperparameter tuning. 
    While a separation into backbones and embedding dimensions provides a fairer comparison, we note some pipeline changes that improve performance independently and should be taken into account when comparing: \red{$^a$}: Better optimizer (RAdam \cite{radam} instead of Adam), \red{$^b$}: Larger input images, \red{$^c$}: Larger batchsizes on SOP, \red{$^d$}: Combination of pooling operations in backbone.}
    \vspace{-5pt}
    \setlength\tabcolsep{1.5pt}
    \footnotesize
    \centering

\resizebox{0.9\textwidth}{!}{
\begin{tabular}{l || c | c | c || c | c | c || c | c | c}
     \toprule
     \multicolumn{1}{l}{\textsc{Benchmarks} $\rightarrow$} & \multicolumn{3}{c}{\textsc{CUB200} \cite{cub200-2011}} & \multicolumn{3}{c}{\textsc{CARS196} \cite{cars196}} & \multicolumn{3}{c}{\textsc{SOP} \cite{lifted}}\\
     \midrule
     \textsc{Methods} $\downarrow$ & R@1 & R@2 & NMI & R@1 & R@2 & NMI & R@1 & R@10 & NMI\\
     \midrule
     \hline
     \multicolumn{10}{>{\columncolor[gray]{.9}}l}{\textbf{ResNet50, 128 dim.}} \\
     \hline
     Margin \cite{margin}  & 63.6 & 74.4 & 69.0 & 79.6 & 86.5 & 69.1 & 72.7 & 86.2 & 90.7\\
     Div\&Conq \cite{Sanakoyeu_2019_CVPR}  & 65.9 & 76.6 & 69.6 & 84.6 & 90.7 & 70.3 & 75.9 & 88.4 & 90.2\\
     MIC \cite{mic}                        & 66.1 & 76.8 & 69.7 & 82.6 & 89.1 & 68.4 & 77.2 & 89.4 & 90.0\\
     PADS \cite{roth2020pads}              & 67.3 & 78.0 & 69.9 & 83.5 & 89.7 & 68.8 & 76.5 & 89.0 & 89.9\\
     S2SD \cite{s2sd} & 68.9 $\pm$ 0.3 & 79.0 $\pm$ 0.3 & 72.1 $\pm$ 0.4 & 87.6 $\pm$ 0.2 & 92.7 $\pm$ 0.2 & 72.3 $\pm$ 0.2& 80.2 $\pm$ 0.2 & 91.5 $\pm$ 0.1 & 90.9 $\pm$ 0.1\\
     \hline
     Multisimilarity+\textit{PLG} & 67.8 $\pm$ 0.2 & 78.2 $\pm$ 0.2 & 70.1 $\pm$ 0.1 & 86.0 $\pm$ 0.3 & 91.4 $\pm$ 0.1 & 72.4 $\pm$ 0.2 & 77.9 $\pm$ 0.1 & 89.9 $\pm$ 0.2 & 90.2 $\pm$ 0.2\\     
     S2SD+\textit{PLG} & \textbf{71.1 $\pm$ 0.1} & \textbf{80.6 $\pm$ 0.2} & \textbf{73.0 $\pm$ 0.2} & \textbf{89.1 $\pm$ 0.2} & \textbf{93.8 $\pm$ 0.2} & \textbf{73.1 $\pm$ 0.3} & \textbf{80.6 $\pm$ 0.1} & \textbf{91.8 $\pm$ 0.2} & \textbf{90.9 $\pm$ 0.1}\\     
     \midrule
     \hline
     \multicolumn{10}{>{\columncolor[gray]{.9}}l}{\textbf{ResNet50, 512 dim.}} \\
     \hline
     EPSHN \cite{epshn}                     & 64.9 & 75.3 &  -   & 82.7 & 89.3 &  -  & 78.3 & 90.7 &  -    \\
     NormSoft \cite{zhai2018classification} & 61.3 & 73.9 &  -   & 84.2 & 90.4 &  -   & 78.2 & 90.6 &  -    \\
     DiVA \cite{milbich2020diva}            & 69.2 & 79.3 & 71.4 & 87.6 & 92.9 & 72.2 & 79.6 & 91.2 & 90.6 \\
     DCML-MDW \cite{Zheng_2021_CVPR_compositional} & 68.4 & 77.9 & 71.8 & 85.2 & 91.8 & 73.9 & 79.8 & 90.8 & 90.8 \\
     IB-DML \cite{seidenschwarz2021graphdml}\red{$^{a,b}$} & 70.3 & 80.3 & \textbf{74.0} & 88.1 & 93.3 & \textbf{74.8} & \textbf{81.4} & 91.3 & \textbf{92.6} \\     
     \hline
     Multisimilarity+\textit{PLG} & 69.6 $\pm$  0.4 & 79.5 $\pm$ 0.2 & 70.7 $\pm$ 0.1 & 87.1 $\pm$ 0.2 & 92.3 $\pm$ 0.3 & 73.0 $\pm$ 0.2 & 79.0 $\pm$ 0.1 & 91.0 $\pm$ 0.1 & 90.0 $\pm$ 0.1\\     
     S2SD+\textit{PLG} & \textbf{71.4 $\pm$ 0.3} & \textbf{81.1 $\pm$ 0.2} & 73.5 $\pm$ 0.3 & \textbf{90.2 $\pm$ 0.3} & \textbf{94.4 $\pm$ 0.2} & 72.4 $\pm$ 0.3 & 81.3 $\pm$ 0.2 & \textbf{92.3 $\pm$ 0.2} & 91.1 $\pm$ 0.2\\     
     \midrule
     \hline
     \multicolumn{10}{>{\columncolor[gray]{.9}}l}{\textbf{Inception-BN, 512 dim.}} \\
     \hline
     Group \cite{elezi2020grouploss} & 65.5 & 77.0 & 69.0 & 85.6 & 91.2 & \textbf{72.7} & 75.1 & 87.5 & 90.8   \\    
     Multisimilarity\red{$^{c}$} \cite{multisimilarity} & 65.7 & 77.0 & - & 84.1 & 90.4 & -  & 78.2 & 90.5 & -   \\
     DR-MS \cite{dutta2020orthogonalunsupdml} & 66.1 & 77.0 & - & 85.0 & 90.5 & -  & - & - & -   \\
     ProxyGML \cite{Zhu2020graphdml} & 66.6 & 77.6 & 69.8 & 85.5 & 91.8 & 72.4  & 78.0 & 90.6 & 90.2   \\
     ProxyAnchor \cite{kim2020proxy}\red{$^{d}$} & 68.4 & 79.2 & - & 86.8 & 91.6 & -  & 79.1 & 90.8 & -   \\     
     \hline
     Multisimilarity+\textit{PLG} & 69.2 $\pm$ 0.2 & 79.7 $\pm$ 0.1 & 70.6 $\pm$ 0.3 & 86.2 $\pm$ 0.2 & 91.5 $\pm$ 0.3 & 70.8 $\pm$ 0.3 & 78.6 $\pm$ 0.2 & 90.7 $\pm$ 0.1 & 90.0 $\pm$ 0.2 \\     
     S2SD+\textit{PLG} & \textbf{70.4 $\pm$ 0.2} & \textbf{80.5 $\pm$ 0.2} & \textbf{71.9 $\pm$ 0.3} & \textbf{88.1 $\pm$ 0.3} & \textbf{92.9 $\pm$ 0.1} & 71.4 $\pm$ 0.3 & \textbf{79.4 $\pm$ 0.1} & \textbf{91.2 $\pm$ 0.1} & \textbf{90.4 $\pm$ 0.2}\\            
     \bottomrule
\end{tabular}}

\vspace{-5pt}
\label{tab:sota}
\end{table*}
\section{Experiments}
\label{sec:experiments}
This section lists experimental details (\S\ref{subsec:experimental_details}, with additional details in \S\ref{supp:sec:exp_details}), highlights the generalization benefits of language guidance (\S\ref{subsec:sota}), and experimentally motivates and ablates the proposed approach in \S\ref{subsec:arch_ablations} - \S\ref{subsec:dense_captions}.

\subsection{Experimental Details}
\label{subsec:experimental_details}
\noindent
\textbf{Datasets.} We provide detailed evaluation of our work on the main DML benchmarks\cite{dvml,daml,hardness-aware,roth2020revisiting,musgrave2020metric,milbich2020diva,s2sd,horde,margin,seidenschwarz2021graphdml}: CUB200-2011 \cite{cub200-2011} (200 bird classes, 11,788 images), Cars196 \cite{cars196} (196 car classes, 16,185 images) and Stanford Online Products \cite{lifted} (22,634 finegrained product classes in 12 supergroups, 120,053 images). Training and test set contain different classes in all benchmarks.\\
\textbf{Implementation Details.} Our experiments use PyTorch \cite{pytorch} with backbones and training protocols adapted from previous research (e.g. \cite{roth2020revisiting,musgrave2020metric,softriple,multisimilarity,milbich2020diva,s2sd}) and codebases following \cite{roth2020revisiting,musgrave2020metric}. 
For our language backbone, we chose the language-part of CLIP \cite{clip} (\textit{ViT-B/32}). However, any big language model can be used instead as well for significant performance improvements, as shown in \S\ref{subsec:arch_ablations}.
For the scaling $\omega$ of our language guidance (see Eq. \ref{eq:final_eq}), we found $\omega\in[1, 10]$ to work consistently for our experiments on CARS196 and CUB200-2011 and $\omega\in[0.1, 1]$ on SOP to account for the magnitude of the base loss $\mathcal{L}_\text{DML}$. For the state-of-the-art study in \S\ref{subsec:sota}, we found these parameter values to transfer well to the other backbones and embedding dimensions. Additional details in Supp. \ref{supp:sec:exp_details}.

\subsection{Benefits of language guidance}
\label{subsec:sota}
We have broken down the literature based on backbone architecture and embedding dimensionality, which are two of the main, DML-independent drivers for generalization performance \cite{roth2020revisiting,s2sd}. 
\footnotetext{We note that our implementation of the Multisimilarity loss follows \cite{multisimilarity} and \cite{musgrave2020metric}, which differs in performance compared to \cite{roth2020revisiting}.}
Stepwise LR-scheduling is performed at most twice based on the performance on a random validation subset (15\%, see e.g. \cite{kim2020proxy,s2sd}).

With little hyperparameter tuning, Tab. \ref{tab:sota} shows competitive and state-of-the-art performance
\footnote{Note: NMI is less reliable, more error prone and dependent on the utilized implementation \cite{musgrave2020metric} than recall-based metrics, which is why beyond comparison to previous literature we primarily utilize Recall@1 and mAP@1000 (measured on recall \cite{roth2020revisiting,musgrave2020metric}).} 
of \textit{PLG}-extended objectives across backbones and embedding dimensions, suggesting that language semantics are very beneficial for visual similarity learning without the need of expert labels. Note that we only compare using \textit{PLG} as \textit{ELG} does not offer any meaningful benefits on SOP due to the absence of finegrained expert classnames (with the exception of the twelve generic superclasses).
Especially for lower-dimensional representation spaces (see ResNet50-128), significant improvements against previous state-of-the-art methods can be seen.
In addition, language guidance boosts base objectives (\textit{Multisimilarity} \cite{multisimilarity}) to match much more complex regularization approaches using e.g. RL-policies \cite{roth2020pads} or joint self-supervised training \cite{milbich2020diva}.

\begin{table}[t]
    \caption{\textit{Relative comparison.} We follow protocols proposed in \cite{roth2020revisiting}\protect\footnotemark, with no learning rate scheduling, to ensure exact comparability. The results show significant improvements when language-guidance is applied. ($^*$) For SOP, only 12 superlabels are given for 11,318 training classes, with very few samples per class. This limits the benefits of language guidance.}
\vspace{-5pt}    
 \footnotesize
  \setlength\tabcolsep{1.4pt}
  \centering
  \resizebox{0.47\textwidth}{!}{
  \begin{tabular}{l || c | c | c}
     \toprule
     \multicolumn{1}{l}{\textsc{Benchmarks}$\rightarrow$} & \multicolumn{1}{c}{\textsc{CUB200}} & \multicolumn{1}{c}{\textsc{CARS196}} & \multicolumn{1}{c}{\textsc{SOP}($^*$)} \\
     \midrule
     \textsc{Approaches} $\downarrow$ & R@1 & R@1 & R@1 \\
    \midrule
    \rowcolor{vvlightgray} 
    \textbf{Multisimilarity} & 62.8 $\pm$ 0.2 & 81.6 $\pm$ 0.3 & 76.0 $\pm$ 0.1\\
    \hline
    +\textit{ELG} & 67.3 $\pm$ 0.2& 85.3 $\pm$ 0.1 & 76.0 $\pm$ 0.2\\ 
    +\textit{PLG} Top-5 & 67.1 $\pm$ 0.4 & 85.4 $\pm$ 0.2 & 76.4 $\pm$ 0.1\\ 
    \midrule 
    \rowcolor{vvlightgray} 
    \textbf{Margin}, $\beta=1.2$ & 62.7 $\pm$ 0.6 & 79.4 $\pm$ 0.5 & 78.0 $\pm$ 0.3\\ 
    \hline
    +\textit{ELG} & 65.3 $\pm$ 0.5 & 83.2 $\pm$ 0.5 & 77.8 $\pm$ 0.1\\ 
    +\textit{PLG} Top-5 & 65.2 $\pm$ 0.5 & 83.4 $\pm$ 0.4 & 78.3 $\pm$ 0.2\\ 
    \midrule 
    \rowcolor{vvlightgray} 
    \textbf{Multisimilarity + S2SD} & 67.7 $\pm$ 0.3 & 86.5 $\pm$ 0.1 & 77.7 $\pm$ 0.2\\ 
    \hline
    +\textit{ELG} & 68.9 $\pm$ 0.4 & 88.2 $\pm$ 0.2 & 77.8 $\pm$ 0.1\\ 
    +\textit{PLG} Top-5 & 69.0 $\pm$ 0.4 & 88.4 $\pm$ 0.3 & 78.0 $\pm$ 0.1\\ 
    \bottomrule 

    \end{tabular}}
    \label{tab:relative_results}
    \vspace{-8pt}
 \end{table}

To ensure that benefits in performance do not solely stem from specific pipeline choices \cite{roth2020revisiting,musgrave2020metric}, we provide additional experimental support following protocols in \cite{roth2020revisiting} with no learning rate scheduling.
As baseline objectives, we select Margin loss with distance-based tuple mining \cite{margin} and the Multisimilarity loss \cite{multisimilarity}, which are among the strongest baseline objectives investigates in \cite{roth2020revisiting} and \cite{musgrave2020metric}. We also apply language guidance alongside state-of-the-art dimensionality-regularization proposed in \cite{s2sd} to showcase benefits even for already heavily regularized objectives. 

Results are highlighted in Tab. \ref{tab:relative_results} (see supplementary for full table with Recall@1, NMI and mAP@1000) for all benchmark datasets. As can be seen, zero-shot generalization performance is improved significantly on CUB200-2011 and CARS196 for both the strong DML baseline objectives (Multisimiliarity \cite{multisimilarity} and Margin loss \cite{margin}, see \cite{roth2020revisiting}) as well as state-of-the-art regularization methods such as S2SD \cite{s2sd}, with improvements in parts of over $4\%$ e.g. for Multisimilarity on CUB200-2011, and nearly $2\%$ for state-of-the-art S2SD on e.g. Cars196 ($86.5\% \rightarrow 88.4\%$). For SOP, we find benefits to be limited - although still significant - both for relative as well as state-of-the-art comparison(e.g. $76.0\% \rightarrow 76.4\%$ relative changes). 
We attribute this to the much higher class-to-sample ratio in SOP (over 11k classes and only few samples per class), resulting in less specific, noisier class alignment.
Unlike pseudo-label guidance however, where performance can be improved reliably, albeit only slightly, we find no benefits in using expert-label guidance. 
This is caused by the absence of finegrained expert class labels - instead, only twelve very coarse and generic superlabels are provided for over 11k classes, with only few samples per class. 
In \textit{PLG}, at least access to the higher resolved 1k ImageNet pseudolabels is given. 


\subsection{Impact of different language context}
\label{subsec:arch_ablations}
\begin{table}[t]
    \caption{\textit{Model guidance quality.} Performance improves regardless of the exact language model and even with large-scale pretrained word embeddings s.a. FastText \cite{fasttext} and GloVe \cite{glove}. However, less transferable word hierarchies fall short in comparison.}
\vspace{-5pt}    
 \footnotesize
  \setlength\tabcolsep{1.4pt}
  \centering
  \resizebox{0.47\textwidth}{!}{
  \begin{tabular}{l || c | c || c | c }
     \toprule
     \multicolumn{1}{l}{\textsc{Benchmarks}$\rightarrow$} & \multicolumn{2}{c}{\textsc{CUB200-2011}} & \multicolumn{2}{c}{\textsc{CARS196}} \\
     \midrule
     \multirow{2}{*}{\textsc{Models} $\downarrow$} & \multirow{2}{*}{R@1} & mAP & \multirow{2}{*}{R@1} & mAP\\
      &  & @1000 & & @1000\\
    \midrule
    \rowcolor{vlightgray}
    \textbf{Baseline} & 62.8 $\pm$ 0.2 & 31.1 $\pm$ 0.3 & 81.6 $\pm$ 0.3 & 31.7 $\pm$ 0.1\\ 
    \rowcolor{vvlightgray}    
    + CLIP-L \cite{clip} & 67.3 $\pm$ 0.2 & \textbf{34.8 $\pm$ 0.2} & 85.3 $\pm$ 0.1 & \textbf{32.7 $\pm$ 0.2} \\
    \midrule        
    \multicolumn{5}{>{\columncolor[gray]{.9}}l}{\textbf{(a) Language Models}} \\
    \hline
    + BERT \cite{bert} & 66.9 $\pm$ 0.3 & 33.5 $\pm$ 0.2 & 84.9 $\pm$ 0.1 & 32.3 $\pm$ 0.1\\
    + Roberta-L \cite{roberta} & 67.3 $\pm$ 0.2 & 33.9 $\pm$ 0.3 & 85.1 $\pm$ 0.2 & 32.4 $\pm$ 0.2\\
    + Reformer \cite{Kitaev2020Reformer} & 66.7 $\pm$ 0.1 & 33.1 $\pm$ 0.1 & 85.5 $\pm$ 0.2 & 32.0 $\pm$ 0.2\\
    + GPT2 \cite{gpt2} & 67.0 $\pm$ 0.3 & 33.7 $\pm$ 0.1 & 84.8 $\pm$ 0.4 & 32.4 $\pm$ 0.1\\
    \hline
    + Top 3 & \textbf{67.5 $\pm$ 0.2} & 34.5 $\pm$ 0.3 & \textbf{85.6 $\pm$ 0.3} & 32.5 $\pm$ 0.3\\   
    \hline
    \multicolumn{5}{>{\columncolor[gray]{.9}}l}{\textbf{(b) Word Embeddings}} \\
    \hline
    + FastText \cite{fasttext} & 66.9 $\pm$ 0.3 & 33.7 $\pm$ 0.3 & 85.2 $\pm$ 0.1 & \textbf{32.7 $\pm$ 0.1}\\
    + GloVe \cite{glove} & 66.1 $\pm$ 0.2 & 33.1 $\pm$ 0.3 & 85.0 $\pm$ 0.2 & 32.1 $\pm$ 0.3\\    
    \hline
    \multicolumn{5}{>{\columncolor[gray]{.9}}l}{\textbf{(c) Word Hierarchies}} \\
    + WordNet \cite{wordnet} & 63.7 $\pm$ 0.2 & 31.3 $\pm$ 0.2 & 82.5 $\pm$ 0.3 & 31.0 $\pm$ 0.2\\
    + HierMatch \cite{hiermatch} & 64.8 $\pm$ 0.3 & 32.2 $\pm$ 0.1 & 82.8 $\pm$ 0.2 & 31.8 $\pm$ 0.1\\
    \bottomrule
    \end{tabular}}
    \label{tab:choice_of_language_models}
    \vspace{-8pt}
 \end{table}
This section studies the impact of different language context on the quality of \textit{ELG}-visual representation spaces (see Table \ref{tab:choice_of_language_models}) with respect to different large language model architectures \textbf{(a)} as well as word embeddings (\textbf{(b)}, FastText \cite{fasttext} and GloVe \cite{glove}\footnote{For classlabels with whitespaces we performed averaging of each respective word embedding.}) trained on large text corpora.

The results show consistent and significant improvements over the vision-only baseline (Tab. \ref{tab:choice_of_language_models}, \textbf{(a)}) for different language model choices s.a. BERT \cite{bert} ($62.8\%$ to $66.9\%$ for Recall@1 on CUB200) or Reformer \cite{Kitaev2020Reformer} ($62.8\%$ to $66.7\%$), as well as for large-scale pretrained word-embeddings (e.g. FastText \cite{fasttext}, $62.8\%$ to $66.9\%$) shown in section \textbf{b}, with little computational overhead ($<5\%$ in walltime) and fast convergence (see Supp. \ref{supp:sec:experiments}). 
This highlights independence of the language model choice, so long as sufficiently general knowledge about semantic class relations is available. We further provide improvements by averaging the top three language models (CLIP-L, Roberta-L and GPT2 - giving $+0.2\%$/$+0.3\%$ on CUB200/CARS196, respectively), but choose CLIP-L for all other experiments to minimize computational overhead. Interestingly, the similar performance of large language models compared to simple, albeit large-scale pretrained, word embeddings indicates that the pretrained language context provided through large language models has yet to be fully utilized for language guidance. We leave this promising direction for future research to investigate further.
 
Finally, section \textbf{c} of Tab. \ref{tab:choice_of_language_models} shows results for guidance based on predefined, but much more specific and less transferable word hierarchies. Specifically, we distill either from Wu-Palmer similarities between class names on WordNet \cite{wordnet} or follow Hierarchical Matching proposed in \cite{hiermatch}. In the latter case, we leverage embeddings computed following \cite{hiermatch} on hierarchy trees derived from wikispecies for CUB200-2011 \cite{hiermatch2} and WordNet otherwise. Training then involves minimizing the similarity between hierarchy and actual image similarities. In both cases, we find that while performance increases somewhat, benefits fall short compared to distillation-based guidance with more general dense language semantics, giving additional support for the benefits of general language knowledge as guidance signal.

\subsection{Conceptual approaches to language inclusion}
\label{subsec:conceptual_ablations}
\begin{table}[t]
    \caption{\textit{Including language context.} Distillation of relative language embedding alignments offers the most benefits.}
\vspace{-5pt}
 \footnotesize
  \setlength\tabcolsep{1.4pt}
  \centering
  \resizebox{0.47\textwidth}{!}{
  \begin{tabular}{l || c | c || c | c }
     \toprule
     \multicolumn{1}{l}{\textsc{Benchmarks}$\rightarrow$} & \multicolumn{2}{c}{\textsc{CUB200-2011}} & \multicolumn{2}{c}{\textsc{CARS196}} \\
     \midrule
     \multirow{2}{*}{\textsc{Methods} $\downarrow$} & \multirow{2}{*}{R@1} & mAP & \multirow{2}{*}{R@1} & mAP \\
     & & @1000 & & @1000 \\
    \midrule
    \rowcolor{vlightgray}    
    Baseline & 62.8 $\pm$ 0.2 & 31.1 $\pm$ 0.2 & 81.6 $\pm$ 0.3 & 31.7 $\pm$ 0.1 \\
    \rowcolor{vvlightgray}    
    + \textit{ELG} & \textbf{67.3 $\pm$ 0.2} & \textbf{34.8 $\pm$ 0.2} & \textbf{85.3 $\pm$ 0.1} & \textbf{32.7 $\pm$ 0.2} \\
    \midrule    
    \multicolumn{5}{>{\columncolor[gray]{.9}}l}{\textbf{(a) Vision-Language representation matching}} \\         
    \hline
    DeVise-Style & 64.2 $\pm$ 0.3 & 32.0 $\pm$ 0.1 & 83.9 $\pm$ 0.3 & 32.2 $\pm$ 0.2 \\    
    ViT Prediction & 65.1 $\pm$ 0.4 & 33.0 $\pm$ 0.1 & 83.2 $\pm$ 0.2 & 32.0 $\pm$ 0.3 \\
    CLIP-style & 63.7 $\pm$ 0.4 & 31.9 $\pm$ 0.6 & 83.2 $\pm$ 0.4 & 32.2 $\pm$ 0.3 \\
    \hline
    \multicolumn{5}{>{\columncolor[gray]{.9}}l}{\textbf{(b) Methods of incorporation}} \\    
    \hline    
    Row-wise L2 & 65.2 $\pm$ 0.2 & 32.4 $\pm$ 0.2 & 84.4 $\pm$ 0.3 & 32.1 $\pm$ 0.1 \\
    Full Similarity & 64.9 $\pm$ 0.3 & 32.3 $\pm$ 0.2 & 84.0 $\pm$ 0.1 & 32.0 $\pm$ 0.3 \\    
    Mining mask & 64.1 $\pm$ 0.3 & 32.6 $\pm$ 0.2 & 81.9 $\pm$ 0.2 & 32.3 $\pm$ 0.4 \\    
    Weight scaling & 59.2 $\pm$ 0.5 & 29.9 $\pm$ 0.4 & 74.6 $\pm$ 0.5 & 26.0 $\pm$ 0.8 \\    
    No stop grad. & 66.3 $\pm$ 0.4 & 34.1 $\pm$ 0.2 & 83.6 $\pm$ 0.2 & 30.8 $\pm$ 0.2 \\      
    \bottomrule
    \end{tabular}}
    \label{tab:architecture_ablations}
    \vspace{-6pt}
 \end{table}

In this section, we motivate structural choice in \textit{ELG} and \textit{PLG} and compare 
against a range of method ablations as well as different approaches for language inclusion in Tab. \ref{tab:architecture_ablations}. See Supp \ref{supp:subsec:concepts} for more details.

More specifically, in section \textbf{a} of Tab. \ref{tab:architecture_ablations}, we investigate different approaches to use language context to shape the visual representation space.
For that, we evaluate training a MLP over image embeddings or a ViT \cite{vit} over sequences of feature vectors (flattened feature maps) to predict language representations by maximizing the cosine similarity between predicted and generated language embeddings (similar to prediction of separately trained word embeddings in DeVise \cite{devise} or caption predictions in multimodal settings \cite{virtex}). Finally, we also look at CLIP-style training as regularizer against $\mathcal{L}_\text{DML}$ by directly contrasting between image and language presentation with no intermediate MLP. In all cases, we found performance to significantly lack behind our distillation-based objective (compare e.g. $63.7\%$ on CUB200 using CLIP-style training to $67.3\%$ using \textit{ELG}), although separately learning a ViT model for language embedding prediction from feature sequence shows some promise for future research ($65.1\%$). We attribute the difference in performance to the fact that our distillation-based matching in \textit{ELG} matches the full similarity matrix of language embeddings without directly interfering with the finegrained visual similarity training.

Finally, section \textbf{(b)} of Tab. \ref{tab:architecture_ablations} looks at different distillation approaches, either matching rows via L2-Distance or the full similarity matrix via KL-Divergence. Additionally, we also include results in which we backpropagate into the language model, as well as leverage language context to directly manipulate the main DML objective $\mathcal{L}_\text{DML}$. For that, we adapt mining and contrastive operations in the Multisimilarity loss (see supp. \ref{supp:subsec:concepts} for exact formulations).
We find that for methods of distillation, relative alignment by row-wise KL-Divergence minimization following Eq. \ref{eq:base_match} works best, motivating the matching approach used in \textit{ELG}. Furthermore, while some benefits where found directly adapting $\mathcal{L}_\text{DML}$ ($62.8\%$ to $64.1\%$ by adjusting mining operations), more hyperparameter tuning is required to achieve reasonable results, while also being objective-specific. 

We similarly also ablated architectural choices in \textit{PLG}, looking at different number of pseudolabels, the application of hierarchical approaches to \textit{PLG} and different methods of distilling from multiple language similarity matrices. However, as results where as expected and in line with experiments conducted in \S\ref{subsec:conceptual_ablations} for \textit{ELG}, we provide the detailed results in Supp. \ref{supp:subsec:plangg_ablations}. In summary however, we found that firstly, even in the pseudolabel setting, hierarchical matching performed notably worse that distillation from large-scale pretrained language models. Secondly, we find a clear benefit of leveraging multiple pseudolabels. Finally, we find not major difference in distillation from a single, averaged pseudolabel similarity matrix versus from multiple pseudolabel-specific similarity matrix. 

\begin{figure}[t]
    \centering
    \includegraphics[width=0.45\textwidth]{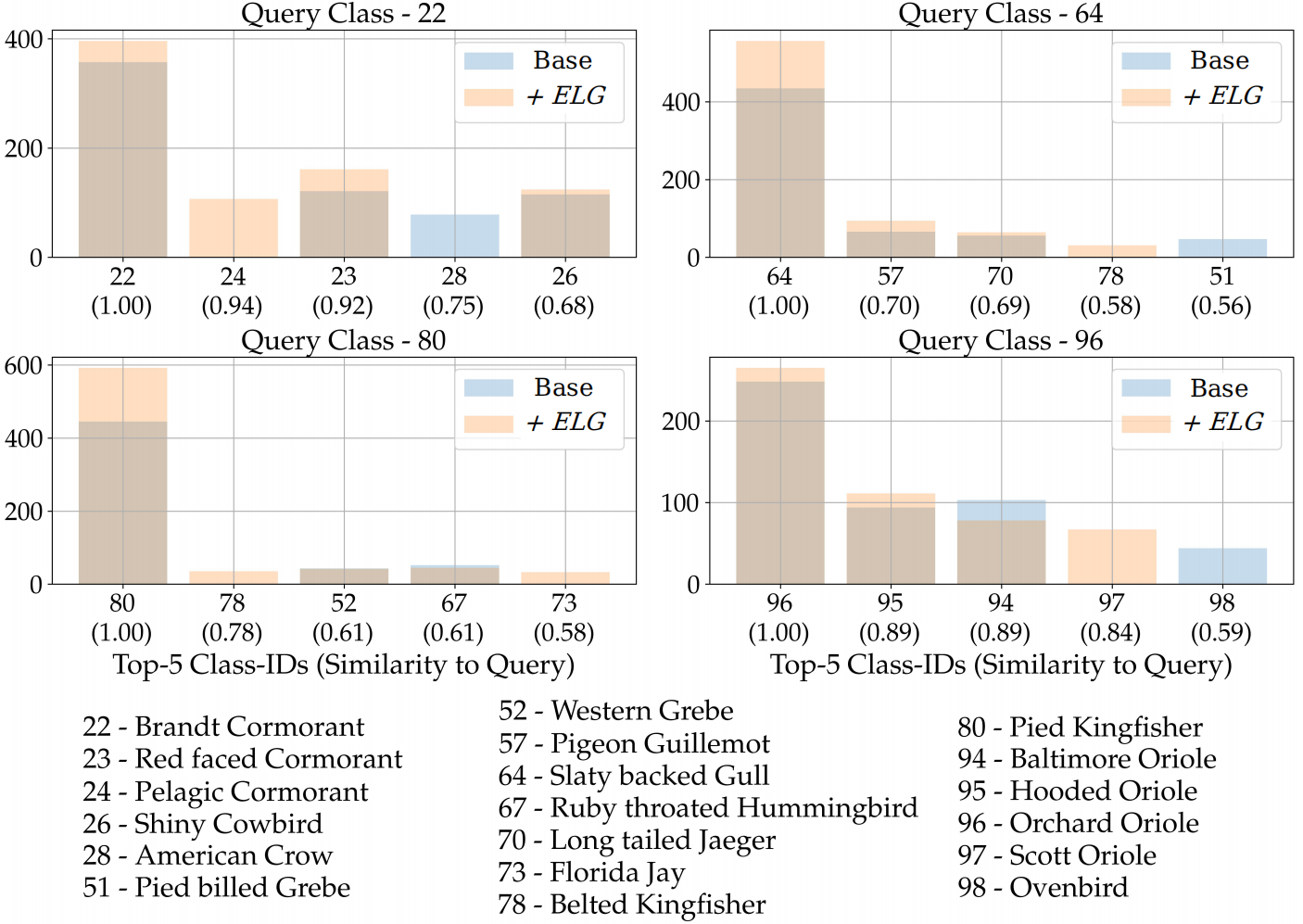}
    \vspace{-5pt}
    \caption{\textit{Improved semantic retrieval.} Embedding test classes on CUB200-2011 for models with and without language guidance and evaluating the Top-5 retrieved classes (sorted by semantic similarity) to a query test classes highlights that language-guided models retrieve more samples from semantically related classes.}
    \label{fig:qualitative_retrieval}
    \vspace{-8pt}
\end{figure}
\subsection{Retrieval based on semantic alignment}
\label{subsec:semantic_alignment}
To understand the representation space impact of language guidance, we embed unseen test classes for models trained with and without. For every class, we use all samples as queries against the available gallery, collect the top twenty nearest retrieved samples and investigate the retrieved classes and respective counts. For both models, we sort the retrieved classes by semantic similarity to the query class (measured by a pretrained language model) and visualize the top five with their respective retrieved counts in Figure \ref{fig:qualitative_retrieval} for exemplary classes on CUB200-2011. 
Indeed, we find that language-guided models retrieve more samples from classes semantically related to the query class. For example, when querying samples from class \texttt{Brandt Cormorant}, methods not equipped with \textit{ELG} fail completely to retrieve samples from the same bird type \texttt{Pelagic Cormorant}, which is heavily semantically related. This is fixed with language guidance and provides clear support that language guidance offers better semantic alignment for visual similarity learning. For methods equipped with \textit{PLG}, we find very similar behaviour.

Finally, we look at the row-wise KL-divergences (see Eq. \ref{eq:base_match}) between full embedding spaces, trained either with or without language guidance, and the actual language embeddings of the respective class names. After application of \textit{ELG}-regularization, we find that alignment more than quadruples (base-to-language divergence of $1.49 \pm 0.16$ versus \textit{ELG}-to-language divergence of $0.34 \pm 0.09$), showcasing the expected realignment based on language semantics.

\subsection{Are dense captions better than class labels?}
\label{subsec:dense_captions}
\begin{table}[t]
\caption{\textit{Do dense captions help?} We find class knowledge to be more valuable than generic image captions for generalization.}
\vspace{-5pt}
\footnotesize
\setlength\tabcolsep{1.4pt}
\centering
\resizebox{0.47\textwidth}{!}{
\begin{tabular}{l || c | c || c | c }
\toprule
\multicolumn{1}{l}{\textsc{Benchmarks}$\rightarrow$} & \multicolumn{2}{c}{\textsc{CUB200-2011}} & \multicolumn{2}{c}{\textsc{CARS196}} \\
\midrule
\multirow{2}{*}{\textsc{Approaches} $\downarrow$} & \multirow{2}{*}{R@1} & mAP & \multirow{2}{*}{R@1} & mAP\\
&  & @1000 & & @1000\\
\midrule
\rowcolor{vlightgray}
Baseline & 62.8 $\pm$ 0.2 & 31.1 $\pm$ 0.2 & 81.6 $\pm$ 0.3 & 31.7 $\pm$ 0.1 \\
\rowcolor{vvlightgray}    
+ \textit{ELG} & 67.3 $\pm$ 0.2 & 34.8 $\pm$ 0.2 & 85.3 $\pm$ 0.1 & 32.7 $\pm$ 0.2 \\
\midrule
Caption & 64.9 $\pm$ 0.4 & 32.2 $\pm$ 0.2 & 84.9 $\pm$ 0.2 & 31.9 $\pm$ 0.3 \\
Caption + insertion & 67.1 $\pm$ 0.2 & 34.3 $\pm$ 0.1 & 85.4 $\pm$ 0.2 & 32.7 $\pm$ 0.3 \\
Classname only & 67.3 $\pm$ 0.3 & 34.7 $\pm$ 0.1 & 85.1 $\pm$ 0.2 & 32.5 $\pm$ 0.1 \\
Other primers & 67.4 $\pm$ 0.1 & 34.7 $\pm$ 0.2 & 85.5 $\pm$ 0.3 & 32.7 $\pm$ 0.1 \\
\bottomrule
\end{tabular}}
\label{tab:caption_ablations}
\vspace{-8pt}
\end{table}

Finally, we check the need for primers (s.a. \texttt{"A photo of a }$y_i$\texttt{"}) and if full image captions can facilitate language guidance.
We try a selection of primer sentences (including class labels only) as well as captions (e.g. \texttt{"A blue bird sitting on a branch."}) generated by a pretrained image caption network (OmniNet \cite{omninet}). We also augment these captions with specific class labels by replacing dataset-dependent keywords (s.a. \texttt{bird}, giving e.g. \texttt{"A blue }$y_i$\texttt{ sitting on a branch"}.
The results in Table \ref{tab:caption_ablations} show that the availability of class knowledge is a stronger guidance proxy than dense image descriptions, which are often too generic (``\textit{Caption}'' vs. \textit{+ELG}). This is supported by the boost in performance when replacing keywords with class labels (``\textit{Caption + insertion}'').
Finally, we find that the exact choice of primer sentence does not matter, with just the classname as input performing only slightly worse than the best primer for each benchmark.


\section{Conclusion}
\label{sec:conclusion}
In this work, we have shown that by matching the relative alignment of visual and language representations, dense language context can help significantly improve the generalization capabilities of visual similarity models.
Extending this approach with ``free'' pseudo-labels, language guidance can be applied on top of arbitrary deep metric learning methods without the need of additional expert knowledge. 
Comprehensive ablation studies and benchmark experiments provide strong support for our proposed method and showcase that language guidance can help existing methods easily achieves competitive and state-of-the-art performance on all standard benchmarks.

\textbf{Limitations.} For language guidance to offer consistent improvements, either expert language classnames or access to suitable pseudolabels has to be available. While ImageNet pretraining is generally transferable to a wide range of downstream tasks, the degree can vary notably. This is exacerbated with high class counts, as seen in the performance on SOP. Additionally, if class separation becomes too finegrained, the likelihood of language models to provide facilitating language context can drop. Finally, for instance-based sample retrieval, where each class comprises samples for a particular class instance, language-guidance is not directly applicable, and requires additional adaptation.

\textbf{Broader Impact.} We investigate re-arranging finegrained visual representation spaces using semantic language context, in parts without the need of external expert labelling. With our experiments highlighting reliable improvements in generalization, this makes our method attractive as a general purpose extension to DML and offer a potential new point of view in how visual metric representation spaces can be learned. As such, there are various applications that can benefit from these insights, such as image or video retrieval as well as general contrastive representation learning. However, we also note that insights in DML also find applications in more controversial areas such as face re-identification. While improvements in DML can thus always be misused, we do not believe that the improvements gained substantially change the societal application. Indeed, we believe that our approach also offers a potential remedy in societal applications, correcting for possible visual feature biases by adjusting to semantic context via respective guidance models.

\section*{Acknowledgements}
This work has been partially funded by the ERC (853489 - DEXIM) and by the DFG (2064/1 – Project number 390727645).
Karsten Roth would like to thank Leonard Salewski (University of Tuebingen) for insightful comments on the use of pretrained language models. 
We thank the International Max Planck Research School for Intelligent Systems (IMPRS-IS) for supporting Karsten Roth.
Karsten Roth further acknowledges
his membership in the European Laboratory for Learning
and Intelligent Systems (ELLIS) PhD program.

{\small
\bibliographystyle{ieee_fullname}
\bibliography{main}
}

\clearpage
\newpage
\begin{centering}
\section*{\Large{Supplementary:  Integrating Language Guidance into Vision-based Deep Metric Learning}}
\end{centering}
\vspace{12pt}

\setcounter{equation}{0}
\setcounter{figure}{0}
\setcounter{table}{0}
\setcounter{page}{1}
\makeatletter
\renewcommand{\theequation}{S\arabic{equation}}
\renewcommand{\thefigure}{S\arabic{figure}}
\renewcommand{\thetable}{S\arabic{table}}

\appendix
\section{Experimental Details}
\label{supp:sec:exp_details}
For all our experiments, we utilize PyTorch \cite{pytorch}. Underlying backbones and training protocols are adapted from previous research (e.g. \cite{roth2020revisiting,musgrave2020metric,softriple,multisimilarity,milbich2020diva,s2sd}) including the codebase provided through \cite{roth2020revisiting}. More specifically, our experiments utilize either a ResNet50 \cite{resnet} with embedding dimensionality of 128 or 512 as well as an Inception-BN \cite{inceptionv1} with dimensionality 512. The ImageNet-pretrained network weights were taken from \texttt{timm} \cite{rw2019timm} as well as \texttt{torchvision} \cite{pytorch}.

Both for studies of relative improvements as well as state-of-the-art performance comparisons, optimization is done using Adam \cite{adam} with a base learning rate of $10^{-5}$, consistent weight decay \cite{weight_decay} of $3\cdot 10^{-4}$ and batchsizes between 80 and 112. Our relative evaluation follows the protocol proposed in \cite{roth2020revisiting}, while for our state-of-the-art comparison, we provide a thorough evaluation against different literature methods separated by the utilized underlying backbone. For our language backbone, we chose the language-part of CLIP \cite{clip} (specifically the ``ViT-B/32'' variant) and the provided tokenizer, but show in section \ref{subsec:arch_ablations} that essentially any big language model can be used for this task. This shows that improvements are not based on potential minor dataset overlap in the image-part of the CLIP training protocol and potential implicit information bleeding into the language model. We also note that the authors of \cite{clip} themselves highlight that even with data overlap, performance is not impacted in a relevant fashion. Applying language-guidance to S2SD \cite{s2sd}, we found placing more emphasis on the feature distillation part instead of the dimensionality distillation worked better in conjunction with both \textit{ELG} and \textit{PLG}. To avoid large-scale hyperparameter grid searches, we thus simply set the weights for dimensionality matching to zero and only adjust the feature distillation weight.

For the scaling $\omega$ of our language-guidance (see Eq. \ref{eq:final_eq}), we found $\omega\in[1, 10]$ to work consistently for our experiments on CARS196 and CUB200-2011 and $\omega\in[0.1, 1]$ on Stanford Online Products, which accounts for the magnitude of the base loss function $\mathcal{L}_\text{DML}$. For the state-of-the-art study in \S\ref{subsec:sota}, we found these parameter values to transfer well to the other backbones and embedding dimensionalities.

\section{Additional Experimental Results}
\label{supp:sec:experiments}
\subsection{Additional language models}
\label{supp:subsec:languagemodels}
To study the impact of language guidance provided with different pretrained language models, Table \ref{supp:tab:choice_of_language_models} provides an extensive evaluation of different language model architectures and pretrainings. As can be seen, performance boosts are consistent, regardless of the exact choice of language model, supporting the general benefit of language as an auxiliary, performance-facilitating modality for finegrained visual similarity tasks.

\subsection{Conceptual approaches to language inclusion}
\label{supp:subsec:concepts}
\begin{table*}[t]
    \caption{\textit{Relative comparison.} We follow protocols proposed in \cite{roth2020revisiting}\protect\footnotemark, with no learning rate scheduling, to ensure exact comparability. The results show significant improvements on CUB200 and CARS196 when language-guidance is applied (with expert- and pseudolabels). ($^*$) For SOP, only 12 superlabels are given for 11,318 training classes. Similarly, SOP only contains very few samples per class, making pseudolabel class estimates very noisy. This makes the benefits of language guidance limited.}
\vspace{-5pt}    
 \footnotesize
  \setlength\tabcolsep{1.4pt}
  \centering
  \resizebox{0.95\textwidth}{!}{
  \begin{tabular}{l || c | c | c || c | c| c || c | c | c}
     \toprule
     \multicolumn{1}{l}{\textsc{Benchmarks}$\rightarrow$} & \multicolumn{3}{c}{\textsc{CUB200-2011}} & \multicolumn{3}{c}{\textsc{CARS196}} & \multicolumn{3}{c}{\textsc{SOP}($^*$)} \\
     \midrule
     \textsc{Approaches} $\downarrow$ & R@1 & NMI & mAP@1000 & R@1 & NMI & mAP@1000 & R@1 & NMI & mAP@1000\\
    \midrule
    \rowcolor{vvlightgray} 
    \textbf{Multisimilarity} & 62.8 $\pm$ 0.2 & 67.8 $\pm$ 0.4 & 31.1 $\pm$ 0.3 & 81.6 $\pm$ 0.3 & 69.6 $\pm$ 0.5 & 31.7 $\pm$ 0.1 & 76.0 $\pm$ 0.1  & 89.4 $\pm$ 0.1  & 43.3 $\pm$ 0.1 \\
    \hline
    +\textit{ELG} & 67.3 $\pm$ 0.2 & 69.6 $\pm$ 0.6 & 34.8 $\pm$ 0.2 & 85.3 $\pm$ 0.1 & 71.7 $\pm$ 0.2 & 32.7 $\pm$ 0.2 & 76.0 $\pm$ 0.2  & 89.5 $\pm$ 0.1  & 43.5 $\pm$ 0.1 \\ 
    +\textit{PLG} Top-5 & 67.1 $\pm$ 0.4 & 69.6 $\pm$ 0.6 & 34.6 $\pm$ 0.5 & 85.4 $\pm$ 0.2 & 71.3 $\pm$ 0.1 & 32.8 $\pm$ 0.2 & 76.4 $\pm$ 0.1  & 89.6 $\pm$ 0.1  & 43.7 $\pm$ 0.1 \\ 
    \midrule 
    \rowcolor{vvlightgray} 
    \textbf{Margin}, $\beta=1.2$ & 62.7 $\pm$ 0.6 & 68.0 $\pm$ 0.3 & 32.2 $\pm$ 0.3 & 79.4 $\pm$ 0.5 & 66.6 $\pm$ 0.7 & 32.8 $\pm$ 0.2 & 78.0 $\pm$ 0.3  & 90.3 $\pm$ 0.2  & 46.3 $\pm$ 0.2 \\ 
    \hline
    +\textit{ELG} & 65.3 $\pm$ 0.5 & 68.5 $\pm$ 0.4 & 33.5 $\pm$ 0.3 & 83.2 $\pm$ 0.5 & 69.0 $\pm$ 0.6 & 33.4 $\pm$ 0.3 & 77.8 $\pm$ 0.1  & 90.2 $\pm$ 0.1  & 46.1 $\pm$ 0.1 \\ 
    +\textit{PLG} Top-5 & 65.2 $\pm$ 0.5 & 68.5 $\pm$ 0.4 & 33.5 $\pm$ 0.3 & 83.4 $\pm$ 0.4 & 69.1 $\pm$ 0.4 & 33.7 $\pm$ 0.3 & 78.3 $\pm$ 0.2  & 90.3 $\pm$ 0.2  & 46.5 $\pm$ 0.1 \\ 
    \midrule 
    \rowcolor{vvlightgray} 
    \textbf{Multisimilarity + S2SD} & 67.7 $\pm$ 0.3 & 71.5 $\pm$ 0.2 & 35.5 $\pm$ 0.2 & 86.5 $\pm$ 0.1 & 71.4 $\pm$ 0.4 & 35.1 $\pm$ 0.3 & 77.7 $\pm$ 0.2  & 89.9 $\pm$ 0.1  & 45.3 $\pm$ 0.3 \\ 
    \hline
    +\textit{ELG} & 68.9 $\pm$ 0.4 & 72.5 $\pm$ 0.3 & 36.4 $\pm$ 0.5 & 88.2 $\pm$ 0.2 & 72.0 $\pm$ 0.1 & 36.0 $\pm$ 0.1 & 77.8 $\pm$ 0.1  & 90.0 $\pm$ 0.2  & 45.3 $\pm$ 0.2 \\ 
    +\textit{PLG} Top-5 & 69.0 $\pm$ 0.4 & 72.4 $\pm$ 0.2 & 36.6 $\pm$ 0.3 & 88.4 $\pm$ 0.3 & 72.4 $\pm$ 0.2 & 36.2 $\pm$ 0.2 & 78.0 $\pm$ 0.1  & 90.0 $\pm$ 0.1  & 45.6 $\pm$ 0.1 \\ 
    \bottomrule 

    \end{tabular}}
    \label{tab:supp_relative_results}
 \end{table*}

\begin{table}[h!]
    \caption{\textit{Models vs. guidance quality.} Performance improves regardless of the exact large pretrained language model. Strong improvements can even be achieved through large-scale pretrained word embeddings such as FastText \cite{fasttext} and GloVe \cite{glove}. However, using less transferable word hierarchies falls short in comparison.}
 \footnotesize
  \setlength\tabcolsep{1.4pt}
  \centering
  \begin{tabular}{l || c | c || c | c }
     \toprule
     \multicolumn{1}{l}{\textsc{Benchmarks}$\rightarrow$} & \multicolumn{2}{c}{\textsc{CUB200-2011}} & \multicolumn{2}{c}{\textsc{CARS196}} \\
     \midrule
     \multirow{2}{*}{\textsc{Models} $\downarrow$} & \multirow{2}{*}{R@1} & mAP & \multirow{2}{*}{R@1} & mAP\\
      &  & @1000 & & @1000\\
    \midrule
    \rowcolor{vlightgray}
    \textbf{Baseline} & 62.8 $\pm$ 0.2 & 31.1 $\pm$ 0.3 & 81.6 $\pm$ 0.3 & 31.7 $\pm$ 0.1\\ 
    \rowcolor{vvlightgray}    
    + CLIP-L \cite{clip} & 67.3 $\pm$ 0.2 & \textbf{34.8 $\pm$ 0.2} & 85.3 $\pm$ 0.1 & \textbf{32.7 $\pm$ 0.2} \\
    \midrule        
    \multicolumn{5}{>{\columncolor[gray]{.9}}l}{\textbf{(a) Language Models}} \\
    \hline
    + BERT \cite{bert} & 66.9 $\pm$ 0.3 & 33.5 $\pm$ 0.2 & 84.9 $\pm$ 0.1 & 32.3 $\pm$ 0.1\\
    + DistBert \cite{distilbert} & 66.7 $\pm$ 0.1 & 33.4 $\pm$ 0.2 & 85.4 $\pm$ 0.4 & 32.4 $\pm$ 0.1\\    
    + Roberta-B \cite{roberta} & 67.0 $\pm$ 0.2 & 33.8 $\pm$ 0.2 & 84.9 $\pm$ 0.1 & 32.3 $\pm$ 0.3\\
    + Roberta-L \cite{roberta} & 67.3 $\pm$ 0.2 & 33.9 $\pm$ 0.3 & 85.1 $\pm$ 0.2 & 32.4 $\pm$ 0.2\\
    + DistRoberta \cite{huggingface} & 66.0 $\pm$ 0.2 & 32.2 $\pm$ 0.2 & 85.0 $\pm$ 0.3 & 32.1 $\pm$ 0.2\\
    + Reformer \cite{Kitaev2020Reformer} & 66.7 $\pm$ 0.1 & 33.1 $\pm$ 0.1 & 85.5 $\pm$ 0.2 & 32.0 $\pm$ 0.2\\
    + MPNet \cite{mpnet} & 66.2 $\pm$ 0.3 & 32.3 $\pm$ 0.2 & 85.4 $\pm$ 0.2 & 32.3 $\pm$ 0.3\\
    + GPT2 \cite{gpt2} & 67.0 $\pm$ 0.3 & 33.7 $\pm$ 0.1 & 84.8 $\pm$ 0.4 & 32.4 $\pm$ 0.1\\
    \hline
    + Top 3 & \textbf{67.5 $\pm$ 0.2} & 34.5 $\pm$ 0.3 & \textbf{85.6 $\pm$ 0.3} & 32.5 $\pm$ 0.3\\   
    \bottomrule
    \end{tabular}
    \label{supp:tab:choice_of_language_models}
 \end{table}
\begin{figure}[h!]
    \centering
    \includegraphics[width=0.48\textwidth]{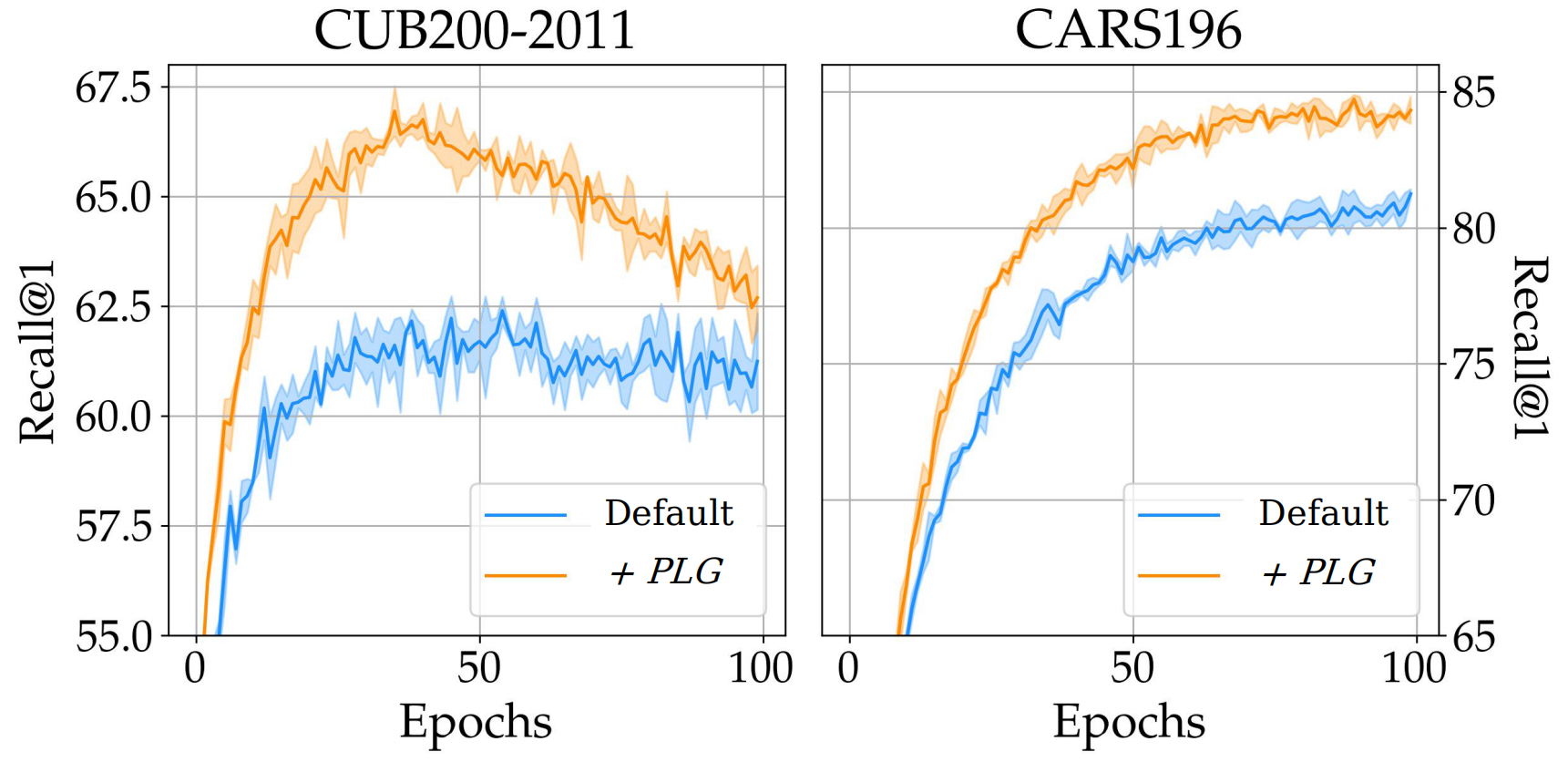}
    \caption{\textit{Impact on convergence.} Matching performance is reached much earlier, with significantly better overall downstream generalization performance.
    }
    \label{fig:convergence}
\end{figure}
To directly incorporate language context into a discriminative DML objective, we utilize the language similarities to either adjust the mining mask or the loss scale in the multisimilarity loss \cite{multisimilarity}. To adjust the mining mask, given an anchor sample $x_a$, positives $x_p$ and negatives $x_n$ are selected if, respectively,
\begin{equation}
\begin{split}
    f\left(S^\text{lang}_{a, p}, s(\psi, \psi_p)\right) &< \min_{\psi_k \in \mathcal{B}, y_k \neq y_a} s(\psi_a, \psi_k) + \epsilon\\
    f\left(S^\text{lang}_{a, n}, s(\psi_a, \psi_n)\right) &< \min_{\psi_k \in \mathcal{B}, y_k = y_a} s(\psi_a, \psi_k) - \epsilon
\end{split}
\end{equation}
with similarity function $s(\bullet, \bullet)$ and language similarity scaling $f(\bullet, \bullet)$. For $f(\bullet, \bullet)$, we investigate different orders of interpolation 
\begin{equation}
f(S^\text{lang}_{ij}, S^\text{img}_{ij})_{\nu_1, \nu_2} = \left[(1 - \nu_1) \cdot {S^\text{lang}_{ij}}^{\nu_2} + \nu_1 \cdot {S^\text{img}_{ij}}^{\nu_2}\right]^{1/{\nu_2}}
\end{equation}
to adjust between sole visual similarity and language similarity.
To re-weight loss components (with positive and negative pairs for anchor $x_a$, $\mathcal{P}^+_a$ and $\mathcal{P}^-_a$), we instead compute
\begin{equation}
\begin{split}
    \mathcal{L}^\textit{ELG}_\text{MSIM} &= \frac{1}{|\mathcal{B}|}\sum_{i=1}^{|\mathcal{B}|} \frac{1}{\alpha}\log\left( 1 + \sum_{k\in\mathcal{P}^+_i} e^{-\alpha\left(\frac{S^\text{lang}_{ik}}{S^\text{img}_{ik}}\right)^{\nu_3}(S^\text{img}_{ik} - \lambda)} \right)\\  
    &+ \frac{1}{\beta} \log\left( 1 + \sum_{k\in\mathcal{P}^-_i} e^{\beta\left(\frac{S^\text{lang}_{ik}}{S^\text{img}_{ik}}\right)^{\nu_4}(S^\text{img}_{ik}-\lambda)} \right)
\end{split}
\end{equation}
thus providing a scaling to the utilized visual similarity $S^\text{img}_{ik}$ based on the (relative) similarity to the respective language similarity.
In all cases, a grid search both over newly introduced hyperparameters ($\nu_1$, $\nu_2$, $\nu_3$ and $\nu_4$) as well as the default multisimilarity loss parameters ($\alpha, \beta, \lambda$) is performed.
For the language similarity scaling $f(\bullet, \bullet)$, we found linear interpolation ($\nu_1 = \nu_2 = 1$) to work best. For $\mathcal{L}^\textit{ELG}_\text{MSIM}$, we found $\nu_3 = \nu_4 = 0.75$ to work well, but had to readjust $\alpha=1.5$ and $\beta=45$ slightly to account for the change in magnitude.

For our matching objective, in which we incorporate language context by training either a MLP over embeddings or a transformer (ViT, \cite{vit}) over a sequence of network features to predict language embeddings $\psi_\text{lang}$, the respective networks are trained following  
\begin{equation}
    \mathcal{L}_\text{Match}(\psi_i, \phi_i, \psi_{\text{lang}, i}) = g_\rho^\text{match}(\psi_i, \phi_i)^T\psi_{\text{lang}, i}
\end{equation}
with $g_\rho^\text{match}(\psi_i, \phi_i)$ denoting the unit-normalized mapping from embedding/feature space to (normalized) language space using either the MLP or ViT with parameters $\rho$.

Finally, as a base reference, we also investigate CLIP-style training in which we utilize direct contrastive training between image and language embeddings following \cite{clip} as regularizer against $\mathcal{L}_\text{DML}$:
\begin{equation}
\begin{split}
    \mathcal{L}_\text{CLIP}(S^\text{mixed}) &= \frac{1}{|\mathcal{B}|}\sum_{i}^{|\mathcal{B}|}-\frac{1}{2}\log\left(\sigma\left(S^\text{mixed}_{i,:}\cdot e^T\right)_i\right)\\ &-\frac{1}{2}\log\left(\sigma\left(S^\text{mixed}_{:,i}\cdot e^T\right)_i\right)
\end{split}
\end{equation}
with similarity matrix between minibatch image and language embeddings $S^\text{mixed}$.

\subsection{Language guidance from pseudolabels}
\label{supp:subsec:plangg_ablations}
\begin{table}[h!]
\caption{\textit{Ablations on Pseudolabel guidance.} More pseudolabels per class improve generalization performance, with class-level pseudolabelling and distillation from a single average language similarity matrix offering highest improvements.}
\footnotesize
\setlength\tabcolsep{1.4pt}
\centering
\begin{tabular}{l || c | c || c | c }
\toprule
\multicolumn{1}{l}{\textsc{Benchmarks}$\rightarrow$} & \multicolumn{2}{c}{\textsc{CUB200-2011}} & \multicolumn{2}{c}{\textsc{CARS196}} \\
\midrule
\multirow{2}{*}{\textsc{Approaches} $\downarrow$} & \multirow{2}{*}{R@1} & mAP & \multirow{2}{*}{R@1} & mAP\\
&  & @1000 & & @1000\\
\midrule
\rowcolor{vlightgray}
Baseline & 62.8 $\pm$ 0.2 & 31.1 $\pm$ 0.2 & 81.6 $\pm$ 0.3 & 31.7 $\pm$ 0.1 \\
\rowcolor{vvlightgray}    
+ \textit{ELG} & 67.3 $\pm$ 0.2 & 34.8 $\pm$ 0.2 & 85.3 $\pm$ 0.1 & 32.7 $\pm$ 0.2 \\
\midrule
\multicolumn{5}{>{\columncolor[gray]{.9}}l}{\textbf{Number of pseudolabels}} \\
\hline
+ \textit{PLG} & 66.2 $\pm$ 0.6 & 33.8 $\pm$ 0.2 & 85.2 $\pm$ 0.1 & 32.7 $\pm$ 0.1 \\    
+ \textit{PLG} (Top-5) & 67.1 $\pm$ 0.4 & 34.6 $\pm$ 0.5 & 85.4 $\pm$ 0.2 & 32.8 $\pm$ 0.2 \\   
+ \textit{PLG} (Top-10) & 67.2 $\pm$ 0.2 & 34.5 $\pm$ 0.3 & 85.3 $\pm$ 0.2 & 32.4 $\pm$ 0.3 \\   
+ \textit{PLG} (Top-20) & 67.0 $\pm$ 0.1 & 34.5 $\pm$ 0.2 & 84.9 $\pm$ 0.4 & 32.1 $\pm$ 0.3 \\   
\midrule
\multicolumn{5}{>{\columncolor[gray]{.9}}l}{\textbf{Word Hierarchies}} \\
\textit{PLG}+WordNet (Top-5) & 64.4 $\pm$ 0.1 & 31.5 $\pm$ 0.2 & 82.9 $\pm$ 0.2 & 31.3 $\pm$ 0.2\\
\textit{pseudo}-HierMatch & 65.0 $\pm$ 0.2 & 32.3 $\pm$ 0.2 & 82.8 $\pm$ 0.3 & 31.6 $\pm$ 0.2\\
\hline
\multicolumn{5}{>{\columncolor[gray]{.9}}l}{\textbf{Different pseudolabel matching methods}} \\
\hline
Sample (Top-5) & 67.0 $\pm$ 0.1 & 34.2 $\pm$ 0.1 & 85.2 $\pm$ 0.2 & 32.2 $\pm$ 0.3 \\   
Dense (Top-5) & 67.0 $\pm$ 0.2 & 33.7 $\pm$ 0.4 & 84.0 $\pm$ 0.1 & 31.5 $\pm$ 0.4 \\   
Multi (Top-5) & 67.2 $\pm$ 0.1 & 34.3 $\pm$ 0.2 & 85.1 $\pm$ 0.2 & 32.2 $\pm$ 0.1 \\   
Dense + Multi & 66.0 $\pm$ 0.3 & 34.1 $\pm$ 0.2 & 84.3 $\pm$ 0.3 & 31.9 $\pm$ 0.2 \\   
\bottomrule
\end{tabular}
\label{tab:plangg_ablations}
\end{table}

For \textit{PLG}, we investigate performance dependence on the number of top-$k$ pseudolabels assigned to every class and their inclusion into training (\S\ref{subsec:language_guided}.
Table \ref{tab:plangg_ablations} highlights that more pseudolabels benefit generalization (optimum for $k\in[5, 10]$), that distillation from a single averaged similarity matrix (see Eq. \ref{eq:multi_match}) performs better than (or comparable to) joint distillation from each pseudolabel similarity matrix (``\textit{Multi (Top-5)}''), and that is does not matter if pseudolabels are computed for classes or individual samples (``\textit{Sample}''). 

In addition, we study whether computing a pseudolabel similarity matrix for each pseudolabel pairing, disregarding the ordering\footnote{E.g. for $k=5$, ``\textit{Dense}'' introduces $k^2 = 25$ target matrices.}, benefits overall performance (``\textit{Dense (Top-5)}'' and ``\textit{Dense + Multi}''), but found no notable benefit. Furthermore, Table \ref{tab:plangg_ablations} shows that leveraging hierarchies as described in \S\ref{subsec:arch_ablations} also performs notable worse in the pseudolabel domain.
Finally, we find impact on overall training time of \textit{PLG} to be negligible, while convergence are in parts even improved (see Supp.-Fig. \ref{fig:convergence}).

\subsection{Convergence of \textit{PLG} models}
Figure \ref{fig:convergence} shows that \textit{PLG (Top-5)} allows underlying objectives to reach similar performance after significantly less training, with much higher overall performance after full training. With \textit{PLG} only requiring an initial forward pass of training samples through the ImageNet-pretrained backbone and of all unique classnames through the language model, impact on overall training time is also negligible.

\end{document}